\documentclass[preprint,11pt]{elsarticle}

\usepackage{graphicx}
\usepackage{float}
\usepackage{amsmath}
\usepackage{booktabs}
\usepackage{algorithm}
\usepackage{algpseudocode}
\usepackage{hyperref}

\usepackage{xcolor}
\usepackage{float}
\usepackage{acronym}
\usepackage{tabularx}
\usepackage{multicol}
\usepackage{array}
\usepackage{makecell}
\usepackage{graphicx}
\usepackage{subcaption}   
\usepackage{caption}
\usepackage{makecell} 
\usepackage{listings}
\usepackage{geometry}
\usepackage{tcolorbox}
\usepackage{lipsum}
\usepackage{amssymb}
\journal{}

\usepackage[utf8]{inputenc}

\usepackage{listings}
\usepackage{xcolor}

\lstdefinelanguage{yaml}{
  keywords={true,false,null,y,n},
  keywordstyle=\color{blue}\bfseries,
  basicstyle=\ttfamily\footnotesize,
  sensitive=false,
  comment=[l]{\#},
  morecomment=[s]{/*}{*/},
  commentstyle=\color{gray}\ttfamily,
  stringstyle=\color{orange},
  showstringspaces=false,
  moredelim=**[il][\color{gray}]{{\#}},
}

\lstdefinelanguage{sam}{
  morekeywords={[1][GlobalParams],[EOS],[Functions],[MaterialProperties],
    [Components],[Preconditioning],[Postprocessors],[Executioner],[Outputs]},
  keywordstyle={[1]\color{blue}\bfseries},
  morekeywords={[2][CH1],[CH2],[CH3],[CH4],[CH5],[pipe1],[pipe2],
    [branch1],[branch2],[inlet_TDV],[outlet_TDV],[reactor]},
  keywordstyle={[2]\color{purple}\bfseries},
  sensitive=false,
  morecomment=[l]{\#},
  commentstyle=\color{gray}\ttfamily,
  stringstyle=\color{orange},
  basicstyle=\ttfamily\scriptsize,
  breaklines=true,
  columns=fullflexible,
  frame=single,
  backgroundcolor=\color{gray!10},
  postbreak=\mbox{\textcolor{red}{$\hookrightarrow$}\space}
}

\begin{document}

\begin{frontmatter}

\title{AutoSAM: an Agentic Framework for Automating Input File Generation for the SAM Code with Multi-Modal Retrieval-Augmented Generation}

% ---- Authors & affiliations ----
\author[nuen,ANL]{Zaid Abulawi\corref{cor1}}
\ead{zaidabulawi@tamu.edu}

\author[nuen,ANL]{Zavier Ndum}

\author[ANL]{Eric Cervi}

\author[ANL]{Rui Hu}

\author[nuen]{Yang Liu}
%\ead{y-liu@tamu.edu}

\address[nuen]{Department of Nuclear Engineering, Texas A\&M University.}
\address[ANL]{Nuclear Science and Engineering Division, Argonne National Laboratory}

% ---- Corresponding authors ----
\cortext[cor1]{Corresponding author.}

\begin{abstract}

In the design and safety analysis of advanced reactor systems, constructing input files for system-level thermal-hydraulics codes such as the System Analysis Module (SAM) code remains a labor-intensive bottleneck. Analysts must manually extract and reconcile design data from heterogeneous engineering documents, then transcribe it into solver-specific syntax. In this paper,  we present AutoSAM, an agentic framework that automates input file generation for SAM. The framework couples a large language model-based agent with retrieval-augmented generation over the solver's user guide and theory manual and specialized tools for analyzing PDFs, images, spreadsheets, and text files. AutoSAM ingests unstructured engineering documents such as system diagrams, design reports, and data tables, extracts simulation-relevant parameters into a human-auditable intermediate representation, and synthesizes validated, solver-compatible input decks. AutoSAM achieves multi-modal retrieval through a pipeline that integrates scientific-text extraction, vision-based figure interpretation, semantic embedding, and query answering, enabling the agent to reason jointly over textual and visual content.

We evaluate the framework on four case studies of progressively increasing complexity: a single-pipe steady-state model, a solid-fuel channel with temperature reactivity feedback, the Advanced Burner Test Reactor core, and the Molten Salt Reactor Experiment primary loop. Across all cases, the agent produces runnable SAM models whose results are consistent with expected thermal-hydraulic behavior, while explicitly identifying missing data and labeling assumed values. The framework achieves 100\% utilization of structured inputs, approximately 88\% extraction from PDF text, and 100\% completeness in vision-based geometric extraction. These results demonstrate a practical path toward prompt-driven reactor modeling, in which analysts provide system descriptions and supporting documentation while the agent translates that intent into transparent, executable, and reproducible SAM simulations. 
\end{abstract}


\end{frontmatter}

\section{Introduction}
\label{sec:intro}

System-level thermal-hydraulics codes are foundational tools in nuclear reactor safety analysis, design evaluation, and regulatory licensing. Codes such as RELAP~\cite{mesina2016history}, TRACE~\cite{bajorek2015development}, CATHARE~\cite{emonot2011cathare}, and the System Analysis Module (SAM)~\cite{hu2025sam} solve one-dimensional conservation equations for mass, momentum, and energy across networks of interconnected components—pipes, heat exchangers, pumps, plena, and reactor core channels. These codes are used to predict system-wide transient behavior under both normal and off-normal conditions~\cite{moshkbar2013transient}. The fidelity of these simulations depends not only on the numerical methods and closure models employed by the solver, but on the correctness and completeness of the input file, which encodes geometry, material properties, boundary and initial conditions, operating states, and solver settings. Input-file errors can propagate into non-conservative safety margins~\cite{d2008best}, making input preparation an important activity in reactor analysis workflows.

Despite the maturity of modern system codes, constructing valid input files remains one of the most labor-intensive and error-prone stages of the modeling and simulation workflow. The process typically requires analysts to extract design parameters from heterogeneous engineering documentation. This includes piping and instrumentation diagrams, design basis documents, material property databases, and legacy reports—and manually transcribe them into solver-specific syntax. This data collection and transcription process is iterative: analysts must reconcile information scattered across text, tables, figures, and spreadsheets, resolve inconsistencies, fill gaps with engineering judgment, and verify that the assembled input file satisfies the solver’s structural and physical constraints. For complex systems such as advanced reactor primary loops, this process can consume days to weeks of analyst effort for a single model~\cite{gupta2001intelligent,macklin2014future}.

The challenge is particularly significant for advanced reactor concepts currently under development, including sodium-cooled fast reactors (SFRs)~\cite{eoh2024design,takano2022core}, molten salt reactors (MSRs)~\cite{csahin2023design,serp2014molten}, fluoride-salt-cooled high-temperature reactors (FHRs)~\cite{scarlat2014design,zweibaum2014phenomenology}, and high-temperature gas-cooled reactors (HTGRs)~\cite{duchnowski2022pre,ueta2011development}. These designs frequently lack the standardized component descriptions and established modeling templates available for conventional light-water reactors. Analysts must therefore construct models largely from heterogeneous sources that may be incomplete, preliminary, or distributed across multiple sources and formats ~\cite{forsberg2018integrated}. 

SAM, developed at Argonne National Laboratory, is a modern system-level simulation code specifically designed to support the thermal-hydraulic analysis of advanced non-light-water reactor concepts~\cite{hu2025sam}. SAM solves the one-dimensional, area-averaged conservation equations and provides models for single-phase and two-phase flow, conjugate heat transfer, and reactor kinetics feedback. The code is actively used in regulatory licensing activities by the U.S. Nuclear Regulatory Commission and supports safety analysis for multiple advanced reactor designs. A SAM input file is organized into distinct parameter blocks—including global parameters, equation of state, components, material properties, functions, executioner, preconditioning, outputs, and postprocessors—each requiring precise specification of parameters drawn from engineering design data. 

The data required to populate these input blocks are often distributed across heterogeneous sources such as text documents, spreadsheets, PDF reports, tables, and images. Extracting simulation-relevant information from these multi-modal documents is typically performed manually, making data collection a time-intensive step that directly affects model fidelity and input file viability. Prior efforts to automate this process have relied on standardized templates, Optical Character Recognition (OCR) \cite{mithe2013optical}, or rule-based key-value extraction. However, these approaches struggle when simulation specifications are fragmented across diverse documents or when essential information, such as boundary conditions, load directions, or coordinate systems, is encoded visually in figures and diagrams. Classical information extraction pipelines may recover isolated facts but frequently fail to map document-level descriptions to solver-specific semantics. Similarly, solver pre-processors and graphical user interfaces \cite{martinez2011graphical} can generate input decks once parameters are explicitly defined, but they do not address the upstream challenge of acquiring those parameters from multi-modal engineering documentation. These limitations motivate the development of a pipeline that jointly interprets multi-modal documents, compiles the extracted information into a structured, solver-specific representation, and enforces validity through systematic verification.

Large language models (LLMs) offer a promising foundation for building such a pipeline, given their demonstrated capabilities in code generation, reasoning, and information synthesis across diverse fields such as material science~\cite{bazgir2025matagent,tang2025matterchat}, biology~\cite{wang2024biorag,fallahpour2025bioreason}, healthcare~\cite{goyal2024healai,yang2024talk2care}, finance~\cite{iacovides2024finllama,wang2025financial}, environmental modeling~\cite{li2024cllmate,hao2025multi}, and engineering applications~\cite{chen2024triz,lin2025fd}. In nuclear engineering, recent studies have begun exploring LLM-enabled automation and intelligent system integration. Ndum et al.~\cite{ndum2025automating} developed an LLM-agent framework to automate Monte Carlo simulation workflows. Liu et al.~\cite{liu2025automating} proposed an LLM-based system for automating data-driven modeling and analysis in engineering applications. Ndum et al.~\cite{ndum2026large} introduced an LLM-assisted digital twin for monitoring and control of advanced reactors. Lim et al.~\cite{lim2025ai} presented an AI-driven thermal-fluid testbed integrating digital twin technologies and LLM interfaces for advanced SMR systems.

However, when applied to specialized domains, LLMs exhibit significant limitations. In fields such as mechanical engineering and materials science, much of the relevant data resides in private databases, secure networks, or within technical figures and tables not readily parsed by language models \cite{xu2023small}. In pharmaceuticals, a large portion of domain-specific knowledge exists within private companies as proprietary or commercially sensitive information \cite{goldacre2017pharmaceutical}. In the nuclear domain, closed-source simulation codes such as SAM are not represented in public training corpora. The absence of such domain-specific data leads to errors or hallucinations when LLMs are prompted outside their knowledge base \cite{tvarovzek2025if}. 

Several strategies have been developed to bridge this gap. Tailored system instructions can guide an LLM to follow specific reasoning patterns and maintain consistency in domain-specific responses \cite{liu2025effects}. However, current models that support very large context windows, exceeding one hundred thousand tokens, face trade-offs when incorporating long instructions, including increased computational cost, slower inference, and reduced effectiveness in capturing long-range dependencies \cite{li2024loogle, liu2024lost}. Retrieval-augmented generation (RAG) provides a complementary approach by granting the model access to external knowledge repositories, retrieving relevant content through similarity search rather than relying solely on the model's training data \cite{xu2023retrieval}. Combining system instructions with RAG can transform a general-purpose LLM into a domain-specific assistant. Beyond knowledge augmentation, equipping LLM-powered agents with specialized tools and libraries enables them to perform complex, multi-step technical workflows that extend well beyond text generation alone.
 
Of particular relevance to nuclear engineering modeling and simulation input file generation, the AutoFLUKA framework \cite{ndum2025automating} demonstrated that a single LLM agent can generate executable Monte Carlo input files for the FLUKA radiation transport code, establishing precedent for AI-assisted simulation input generation in the nuclear domain. However, existing approaches have generally focused on single-source, text-based inputs and have not addressed the multi-modal document processing challenge inherent in system-code modeling, where critical information is distributed across PDFs, engineering drawings, images, spreadsheets, and narrative reports. Because SAM is closed-source and licensed, its input syntax, parameter definitions, and modeling conventions are not well represented in the training data of current LLMs, further limiting the direct applicability of general-purpose AI assistants to SAM-related tasks.

In this work, we present AutoSAM, a multi-modal, retrieval-augmented LLM agent framework designed to automate the generation of SAM input files directly from heterogeneous engineering documentation. The framework employs a ReAct-based single agent architecture that combines three complementary specialization strategies: (1) expert-engineered system instructions encoding SAM’s input file structure and modeling conventions, (2) retrieval-augmented generation over embedded representations of SAM’s user guide and theory manual, and (3) a suite of seven specialized tools for parsing PDFs, images, spreadsheets, and text files, as well as for creating and validating SAM input decks. The agent ingests unstructured engineering artifacts, extracts simulation-relevant parameters into a human-auditable intermediate structured representation, and synthesizes validated, solver-compatible input files. The intermediate representation serves as a deliberate human-in-the-loop checkpoint, allowing domain experts to review, correct, and augment the extracted data before final input generation—a design choice aligned with the quality assurance expectations of nuclear safety analysis.

We demonstrate and evaluate the framework through four case studies of progressively increasing complexity: (i) a single-pipe steady-state model that validates baseline component modeling, (ii) a solid-fuel channel with temperature-driven reactivity feedback that tests multi-physics coupling, (iii) the reactor core of the Advanced Burner Test Reactor (ABTR) that requires multi-modal document processing for a parallel-channel configuration, and (iv) the primary loop of the Molten Salt Reactor Experiment (MSRE) that challenges the agent to reconstruct full-loop topology and connectivity from heterogeneous sources.

The remainder of this paper is organized as follows. Section \ref{sec:methodology} presents our methodology, detailing the personalization of the ReAct agent through system instructions, RAG, and the integration of specialized tools. Section \ref{sec:tools} discusses the customized tool we developed for AutoSAM to retrive and process multi-modal knowledge. Section \ref{sec:results} provides the test cases and the results. Section \ref{sec:discuss} discusses personalization of the methodology, model limitations, and safety considerations. Finally, Section \ref{sec:conclusions} concludes with an assessment of the strengths and limitations of the current work.

\section{Methodology}
\label{sec:methodology}

The limited performance of current LLMs on physics and engineering solvers can primarily be attributed to the scarcity of data describing solver theory, usage, and input syntax. In some cases, the closed-source nature of the code further restricts the availability of training data. This limitation reflects a lack of exposure to relevant domain-specific information rather than an inherent inadequacy in LLM capabilities. To develop an intelligent agent capable of assisting users in generating input files for such solvers, the agent must be specialized and equipped with foundational knowledge of the target code. In the present work, this domain knowledge is introduced through three complementary strategies: (1) tailored system instructions, (2) retrieval-augmented generation (RAG), which grants the model access to external, context-specific documents during inference, and (3) specialized tools aligned with the target domain. The framework is implemented using LangChain, which provides the orchestration needed to integrate the LLM, retrieval pipeline, and domain-specific tools into an agentic workflow.

\begin{figure}[H]
    \centering
\includegraphics[width=1\textwidth]{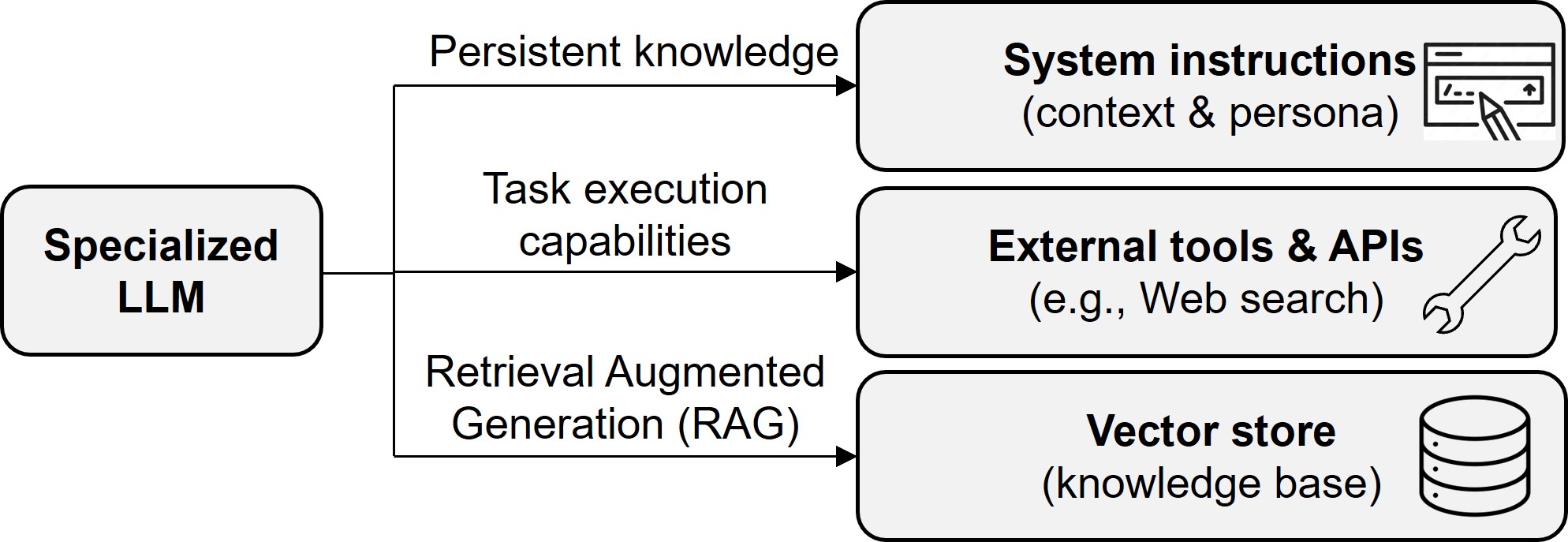} % Change to your file name and extension
    \caption{Specialized LLM-agents through tools, RAG, and system instructions.}
    \label{fig:llm_rag_tool}
\end{figure}

Figure~\ref{fig:llm_rag_tool} illustrates this three-pillar specialization approach.

The following subsections first describe the SAM system code and its input file structure, which define the domain that the agent must support. We then present how domain-specific knowledge is delivered through system instructions and RAG, the document processing workflow that governs how engineering documents are converted into solver-ready inputs, and the three categories of tools that enable the agent to execute this workflow.

\subsection{System Analysis Module code}
\label{sec:sam}

\begin{figure}[H]
    \centering
\includegraphics[width=1\textwidth]{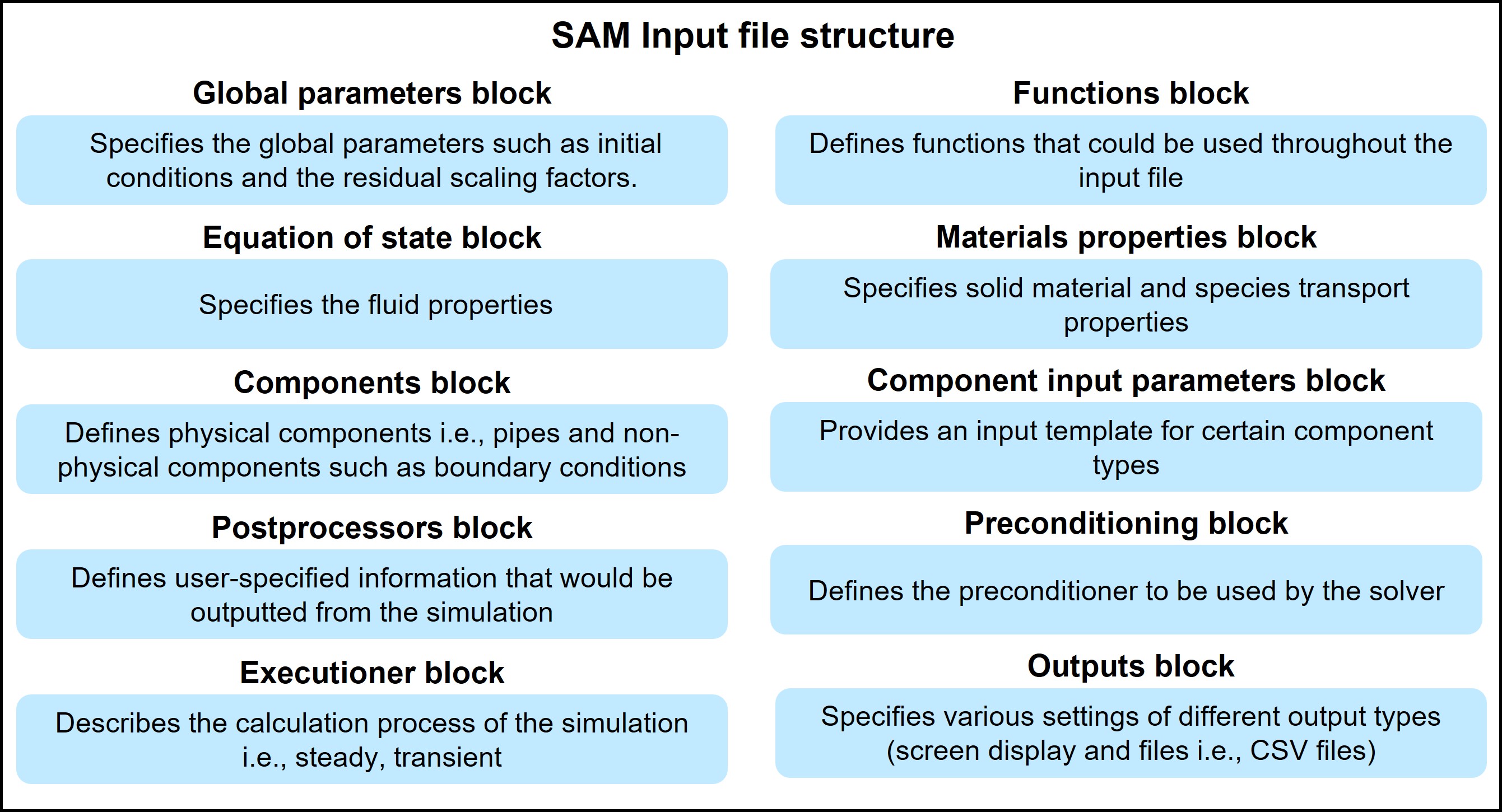} % Change to your file name and extension
    \caption{SAM system code input file structure.}
    \label{fig:input_file}
\end{figure}

SAM is a  system-level simulation code developed by Argonne National Laboratory to support the design and safety evaluation of advanced non-light water reactor systems. SAM provides a framework capable of modeling the complex thermal-hydraulic behavior of these reactors under both normal and off-normal conditions. Despite its focus on one-dimensional (1-D) simulations, SAM incorporates models for heat transfer, fluid dynamics, and system interactions. A SAM input file defines the configuration and behavior of a system-level simulation through a collection of parameter blocks, each with a distinct purpose in the modeling process, as shown in Figure\ref{fig:input_file}. 

The Global Parameters block initializes the system with values such as the initial temperature and pressure. The equation of state block specifies the thermodynamic properties of the coolant, which govern fluid behavior. The components block is used to define the physical system of pipes, heat exchangers, pumps, etc. It also provides the boundary conditions and defines the connection between different components. The Materials Properties block defines solid material characteristics, including thermal conductivity, specific heat, and density for components. The Functions block allows the user to define mathematical expressions that could be used throughout the model. The simulation’s numerical configuration is specified in the Executioner block, which sets the mode of operation (e.g., transient or steady), time discretization, and convergence criteria. The Preconditioning block provides additional solver control.The outputs block defines how the data are exported, whether to screen, excel files or other options. The Postprocessors block allows users to extract derived quantities, or compute location-specific metrics like temperature or pressure. The limited performance of current LLMs on the SAM system code can primarily be attributed to the closed-source nature of the code, which restricts the availability of training data.

\subsection{Specific Domain Knowledge through System Instructions and RAG}

Domain-specific knowledge is incorporated into the agent through two complementary approaches: system instructions and RAG. First, foundational knowledge is delivered via system instructions—a persistent context that is always visible to the model regardless of the user’s prompt. These instructions are carefully engineered and curated from the solver user guide and theory manual. Due to limitations imposed by the model's context window, only the most essential and foundational information is included. The system instructions define the agent's role and personalize its behavior, specifying the types of tasks it should support, such as writing input files, assisting in data collection, validating the input file, and troubleshooting errors and warnings to correct input files. The instructions also outline the relevant domains the agent should focus on, including fluid flow, heat transfer, thermodynamics, and reactor physics. In addition, basic information about the input file structure is provided, including required blocks and parameters. The instruction set is expandable based on the agent’s observed performance and includes few shots of the solver input files. Second, supplementary knowledge is made accessible through RAG. The solver's user guide and theory manual are embedded into a vector store, enabling the agent to retrieve relevant information through similarity search based on the user's prompt. 

\subsection{Documents processing workflow}

The automated input generation workflow begins with the ingestion of unstructured user-provided documents such as images, and PDF files that contain model description, geometry and topology specifications, materials properties, and other simulation related parameters. The structured documents  processed using specialized tools, including a PDF Analysis Tool for extracting text, tables, equations, and PDF images, an Image Analysis Tool for interpreting schematics, designs, and layouts. The extracted information is then structured and passed into an LLM. The agent produces an intermediate structured YAML file that captures all necessary simulation data in a modular, human-readable format. This intermediate file is not yet the solver syntax input file, but it provides a flexible, organized representation of the simulation setup.

\begin{figure}[t]
    \centering
\includegraphics[width=1\textwidth]{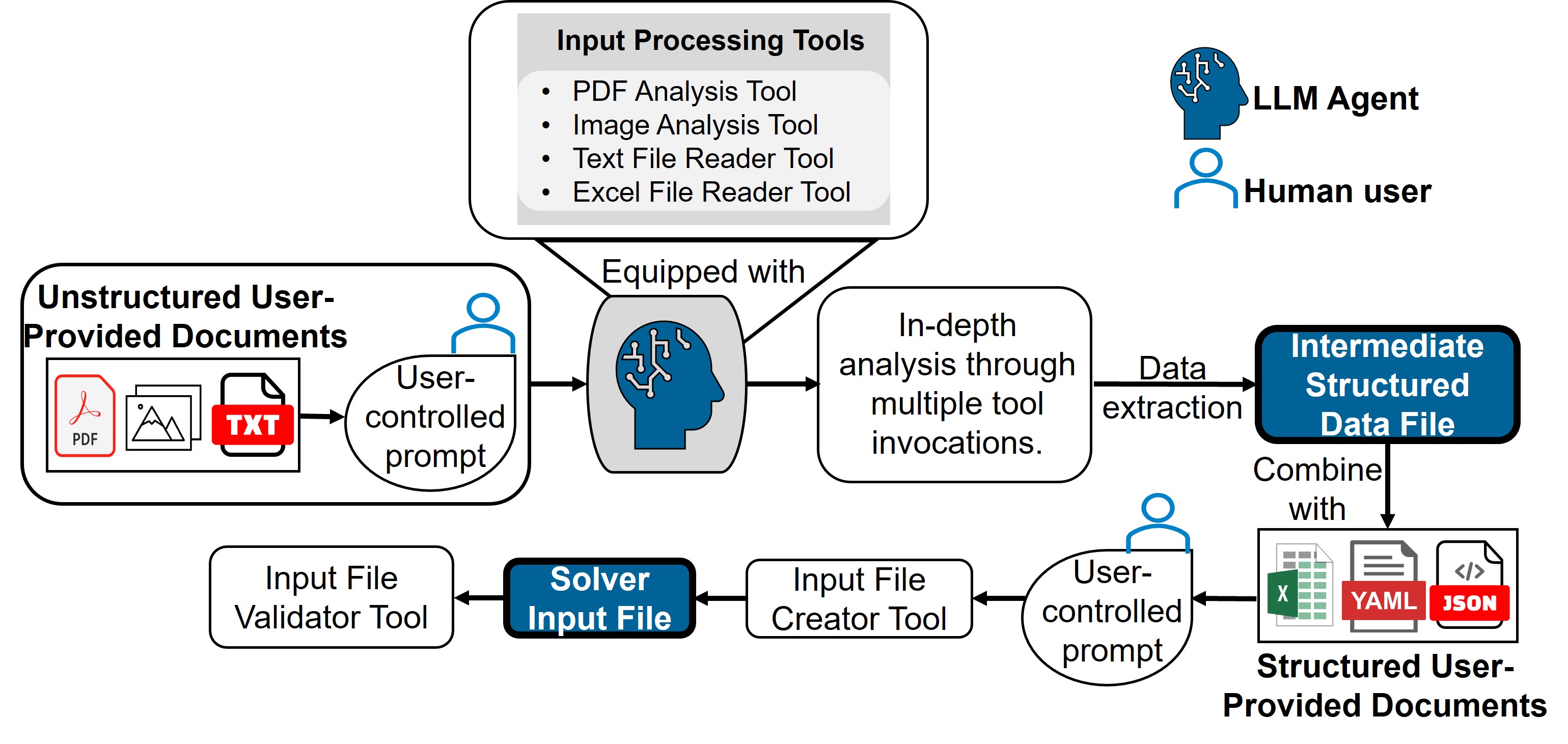} % Change to your file name and extension
    \caption{A streamline of the agents tools for multi-modal documents processing using a user dynamic prompt.}
    \label{fig:workflow}
\end{figure}

The intermediate file serves a crucial role as a human-in-the-loop checkpoint between raw document data and the final solver input file. It allows domain experts to verify, edit, and augment the model before code generation, which is vital for correcting possible misinterpretations or making nuanced decisions that an LLM alone might miss. The intermediate structured file is then combined with other user-provided structured documents such as Excel, JSON, and YAML files. Then the agent transforms the data in the structured documents into a solver-compatible input file by the Input Creator Tool, and an Input File Validation Tool is used to ensure completeness and consistency. The document processing workflow is illustrated in Figure~\ref{fig:workflow}.

\section{Tools for AutoSAM}
\label{sec:tools}
\begin{figure}[H]
    \centering
\includegraphics[width=0.95\textwidth]{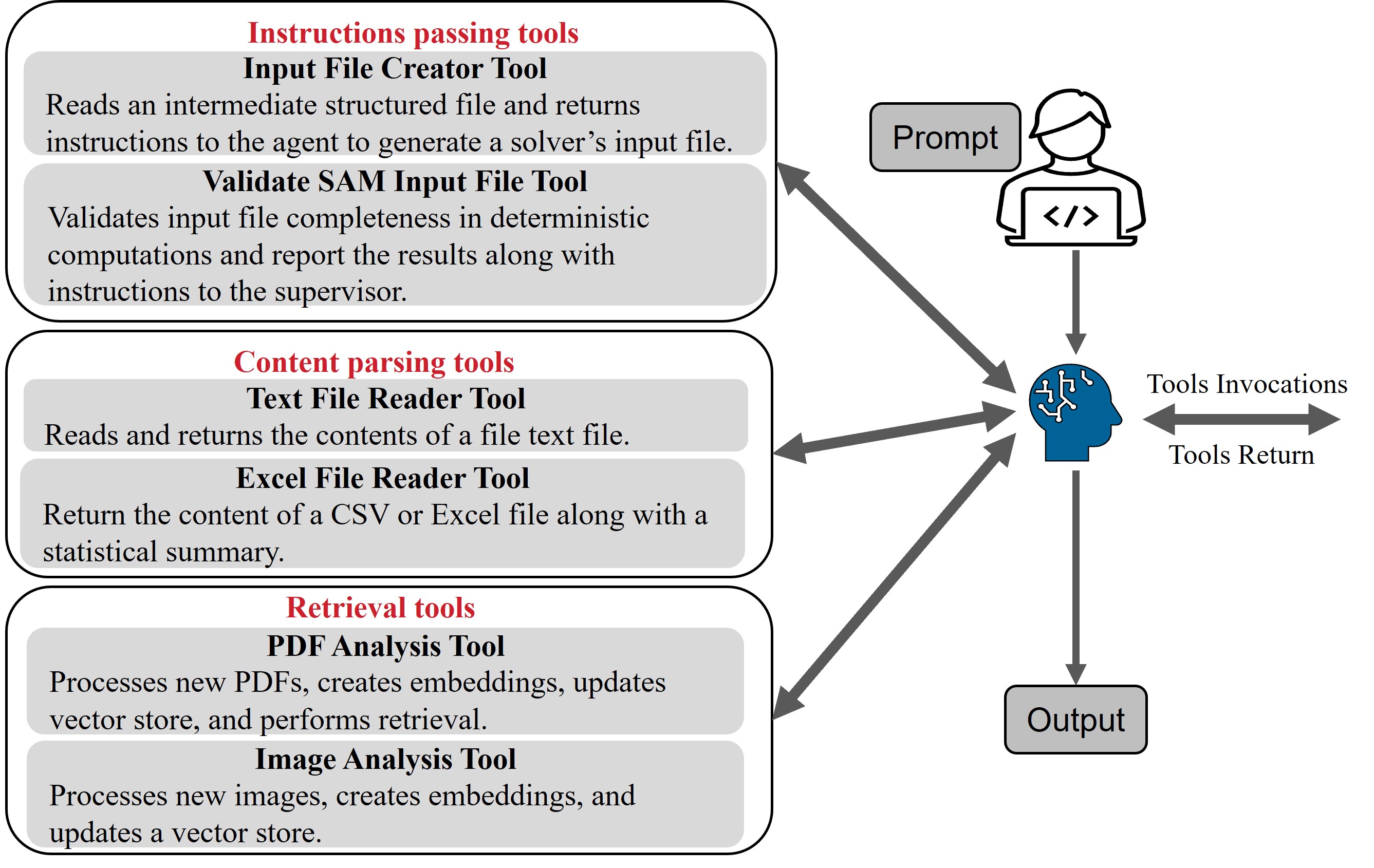} % Change to your file name and extension
    \caption{A summary of the tools' capabilities: content parsing, instruction passing, retrieval, and Python code execution tool.}
    \label{fig:tools}
\end{figure}

For the agent to support the workflow, and the data collection through documents processing, tools are needed. Some of these tools are essential to perform the data collection, while others are performance-enhancing. There are seven tools in total, divided into two content parsing tools, two instruction passing tools, two retrieval tools, and a Python code execution tool. The tools are summarized in Figure~\ref{fig:tools}. In this setup the agent could invoke a tool or multiple tools in a single call based on the user prompt and documents provided. After the tools are executed, a tool return is sent to the agent, where further actions and a response could be generated.

\subsection{Content parsing tools}

Content parsing tools return the entire file content to the agent due to the nature of such files. For example, Excel files may contain tabulated design data where all the information is required to generate an input file, making direct parsing the best option. Similarly, a user may provide an incomplete input text file whose full content must be parsed before the agent can complete it.

\begin{figure}[H]
    \centering
\includegraphics[width=0.48\textwidth]{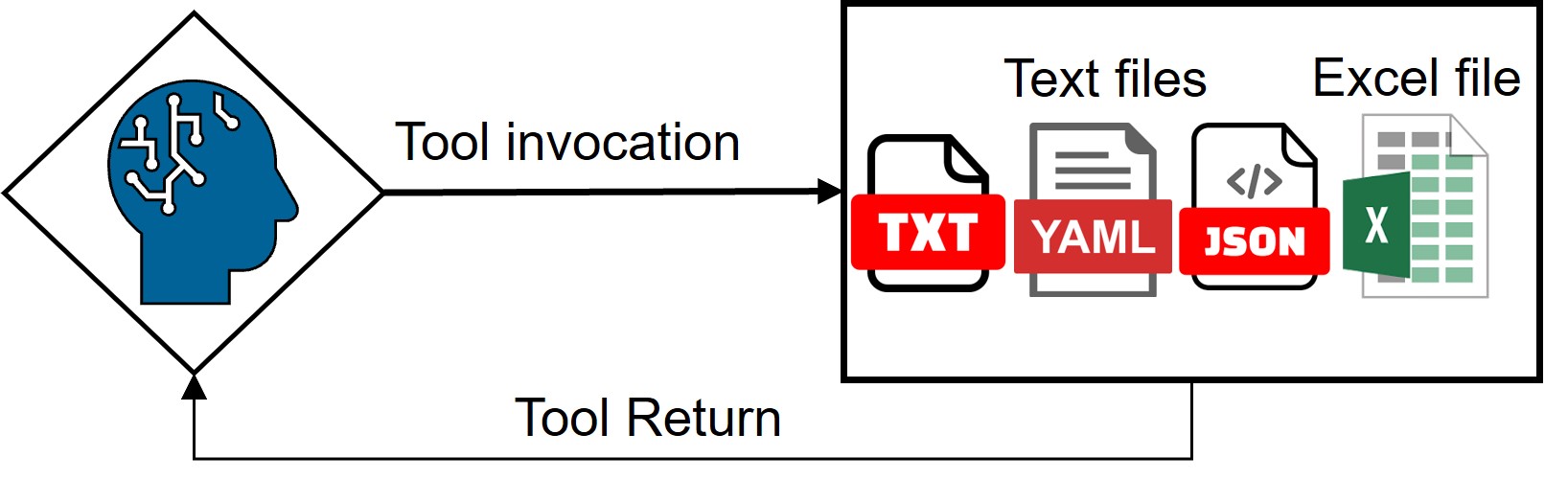} % Change to your file name and extension
    \caption{Depiction of the text and Excel files content parsing tool.}
    \label{fig:reader}
\end{figure}

The Excel file reader tool is designed to automatically read, summarize, and provide insights from tabular data files. This summary includes the list of column names, the file data, descriptive statistics, and structural metadata. The Text file reader tool is designed to read and return the contents of various structured and unstructured text-based files. Figure~\ref{fig:reader} depict the content parsing tools. 

\subsection{Retrieval tools}
The retrieval tools are meant to create embeddings of PDF files and images, and then perform retrieval based on a user query. %These tools are designed in a way that if the embedding is already created, the tool will move to the retrieval directly.
The PDF analysis tool is a multi-stage pipeline designed to extract the contents, perform semantic embedding, and retrieve scientific knowledge from PDF documents. It leverages vision-language models to extract all the PDF document information including text, equations, tables, and figures. The tool performs four main operations: (1) structured text extraction using Meta AI's Nougat model (Neural Optical Understanding for Academic Text), (2) figures' caption and description generation via vision LLMs, (3) vector embedding of the extracted content, and (4) retrieval-augmented query answering. 

\begin{figure}[H]
    \centering
\includegraphics[width=1\textwidth]{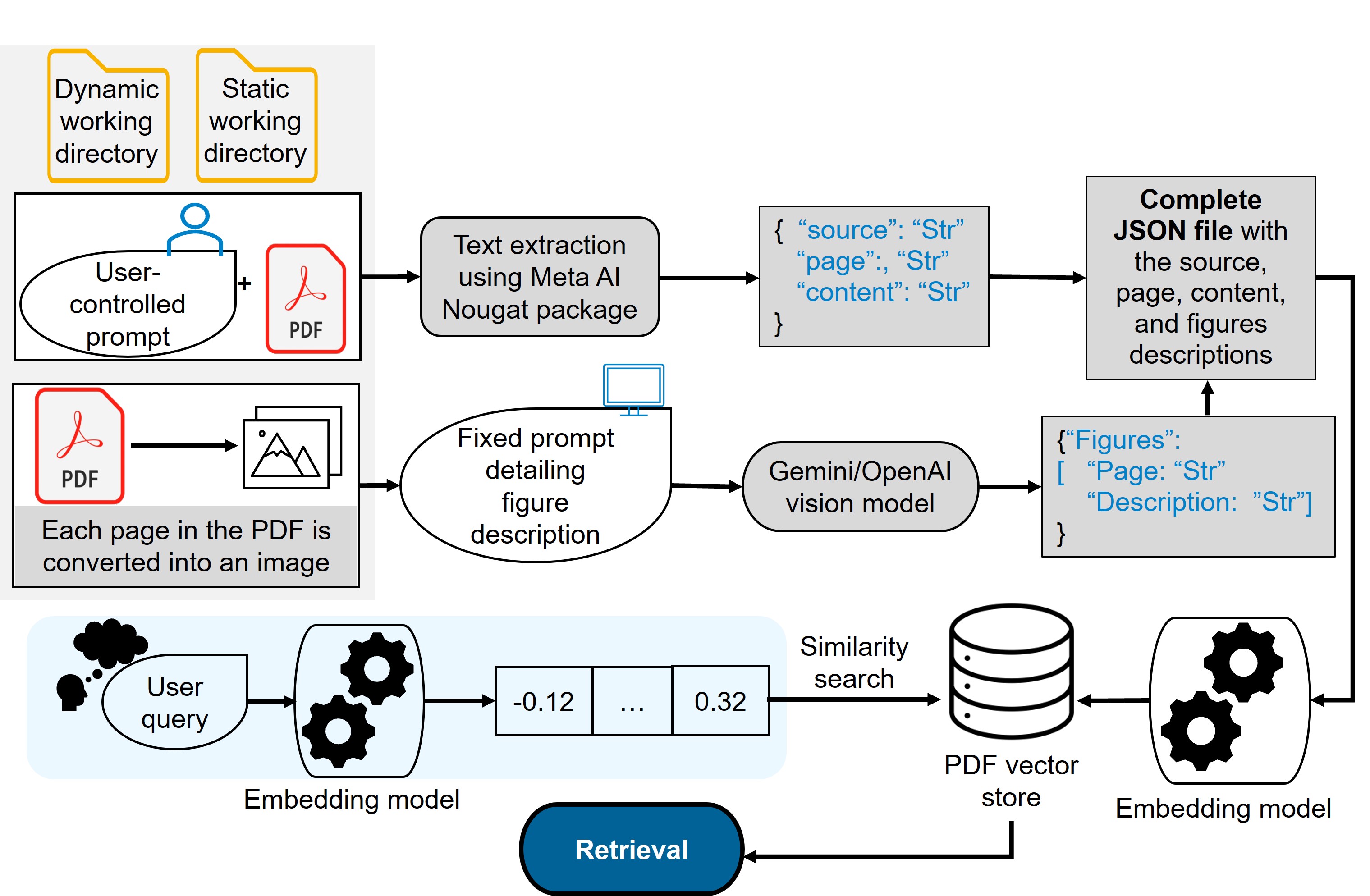} % Change to your file name and extension
    \caption{A depiction PDF analysis tool embedding and retrieval logic.}
    \label{fig:pdf_tool}
\end{figure}

In the first stage, PDF documents are processed using the Nougat model, a vision-to-text system trained specifically on scientific papers. Unlike conventional OCR, Nougat operates directly on full-page images, reconstructing the original layout—including paragraphs, section headers, tables, and even complex LaTeX-style mathematical expressions. The extracted text is then semantically chunked and encoded as a dictionary with metadata such as the PDF file name, page number, and page-level content. This representation allows the system to reference specific documents and pages when responding to user queries. Nougat model isn't able to extract information from images inside the PDF document, thus further handling is needed. 

After text extraction, each page is rasterized into a high-resolution PNG image. These images are analyzed by a vision-language model using a structured prompt instructing it to detect and describe all visual figures on the page, including diagrams, plots, schematics, and other scientific illustrations. The resulting figure descriptions are stored in a centralized knowledge index, enriching the document with visual semantic information.

figure metadata is added to the extracted data from the Nougat model, then the tool computes vector embeddings for all text and figure chunks. These embeddings, which support similarity-based retrieval, are stored for downstream use. In the final stage, a user query is embedded and matched against the stored vectors. The top-retrieved results are provided to the agent, which generates a human-readable answer and cites the most relevant source pages. The output is returned as a Markdown-formatted response that includes both the explanation and its references. This workflow supports integrated text-based and figure-based retrieval, enabling the system to reason over both visual and linguistic content in scientific PDF documents. Figure~\ref{fig:pdf_tool} provides a visual overview of the PDF analysis pipeline.

In addition to retrieving from a dynamic directory with user-provided documents, the PDF retrieval tool is able to retrieve data from a static directory. This setup combines long-term shared knowledge with user-specific content, where the static directory provides information on writing and constructing an input file, and a dynamic directory provides input data required to build the model. 

\begin{itemize}
    \renewcommand\labelitemi{--}
    \item \textbf{Static Directory:} A permanent folder containing a prebuilt vector store of shared domain knowledge, typically created from core documents such as theory manuals and user guides.
    \item \textbf{Dynamic Directory:} A user-provided working directory containing its own vector store, built from newly added or task-specific documents.
\end{itemize}

\begin{figure}[H]
    \centering
\includegraphics[width=1\textwidth]{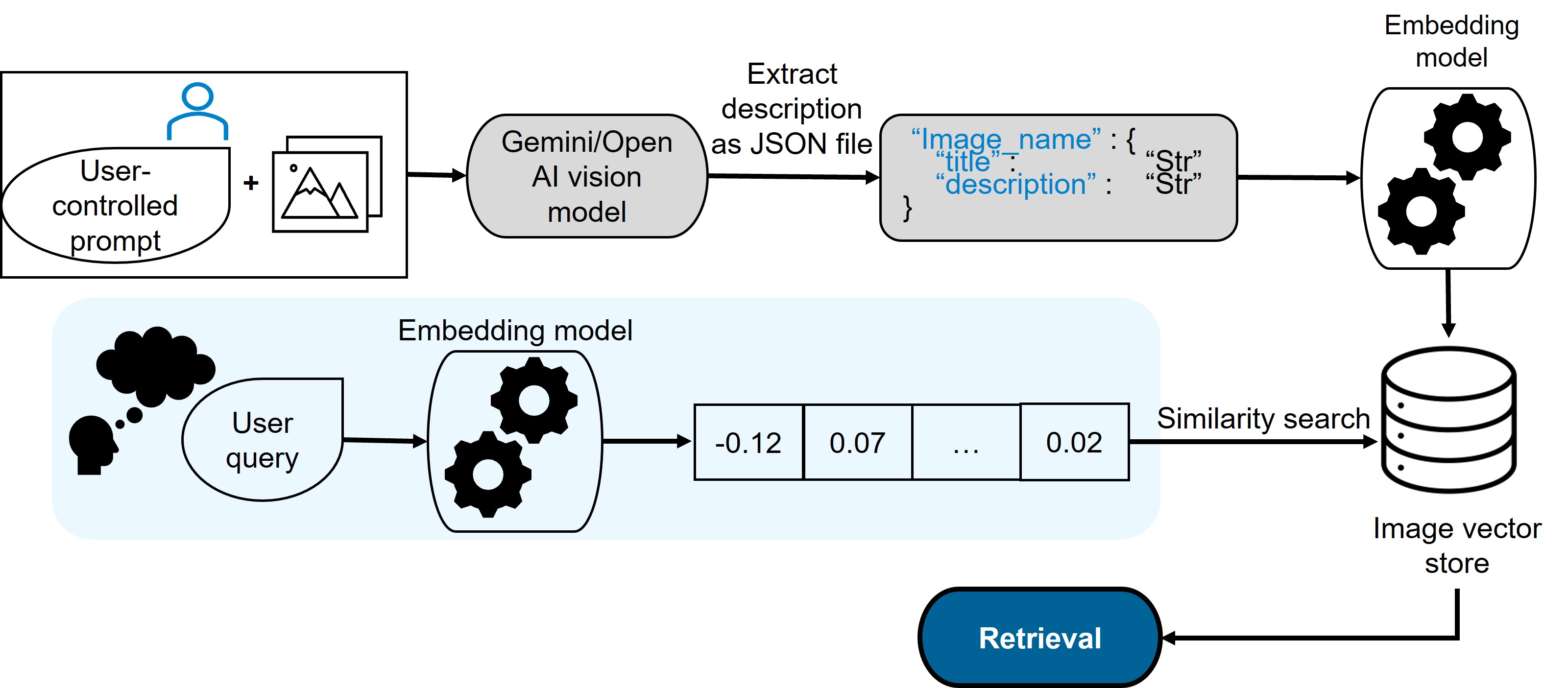} % Change to your file name and extension
    \caption{A depiction of the image analysis tool embedding and retrieval logic.}
    \label{fig:image_tool}
\end{figure}

The Image analysis tool enables automated understanding and retrieval of images using vision-language models and vector embeddings. It is designed to scan a folder of images, describe each new image using a user-controlled prompt with an LLM, embed the image descriptions semantically, and answer user queries through similarity-based retrieval. This makes it ideal for knowledge-driven systems that require understanding visual representations such as plots, schematics, or experiment screenshots. Figure~\ref{fig:image_tool} illustrates the full workflow of the image analysis tool. For each new image not already processed, the tool sends the image and user-defined prompt to a selected vision-language model, extracts a detailed natural language description of the image, records the description and title into a structured file, and embeds the description using a vector embedding model. If a user query is provided, the tool semantically retrieves the most relevant image descriptions from the vector store using similarity matching. 

\subsection{Instruction passing tools}
Instructions passing tools return a set of instruction for the agent to complete a certain task. These instructions are expandable based on the agent's performance. 

\begin{figure}[H]
    \centering
    \begin{subfigure}[b]{0.48\textwidth}
        \centering
        \includegraphics[width=\textwidth]{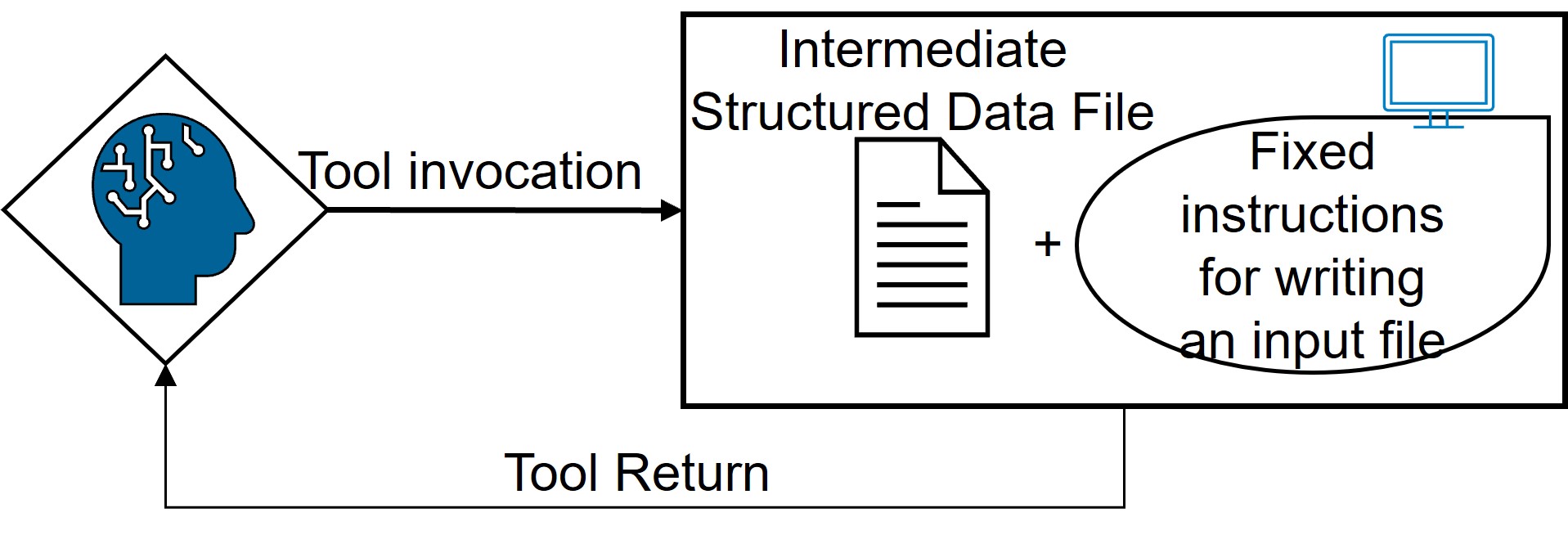}
        \caption*{(a)}
    \end{subfigure}
    \hfill
    \begin{subfigure}[b]{0.48\textwidth}
        \centering
        \includegraphics[width=\textwidth]{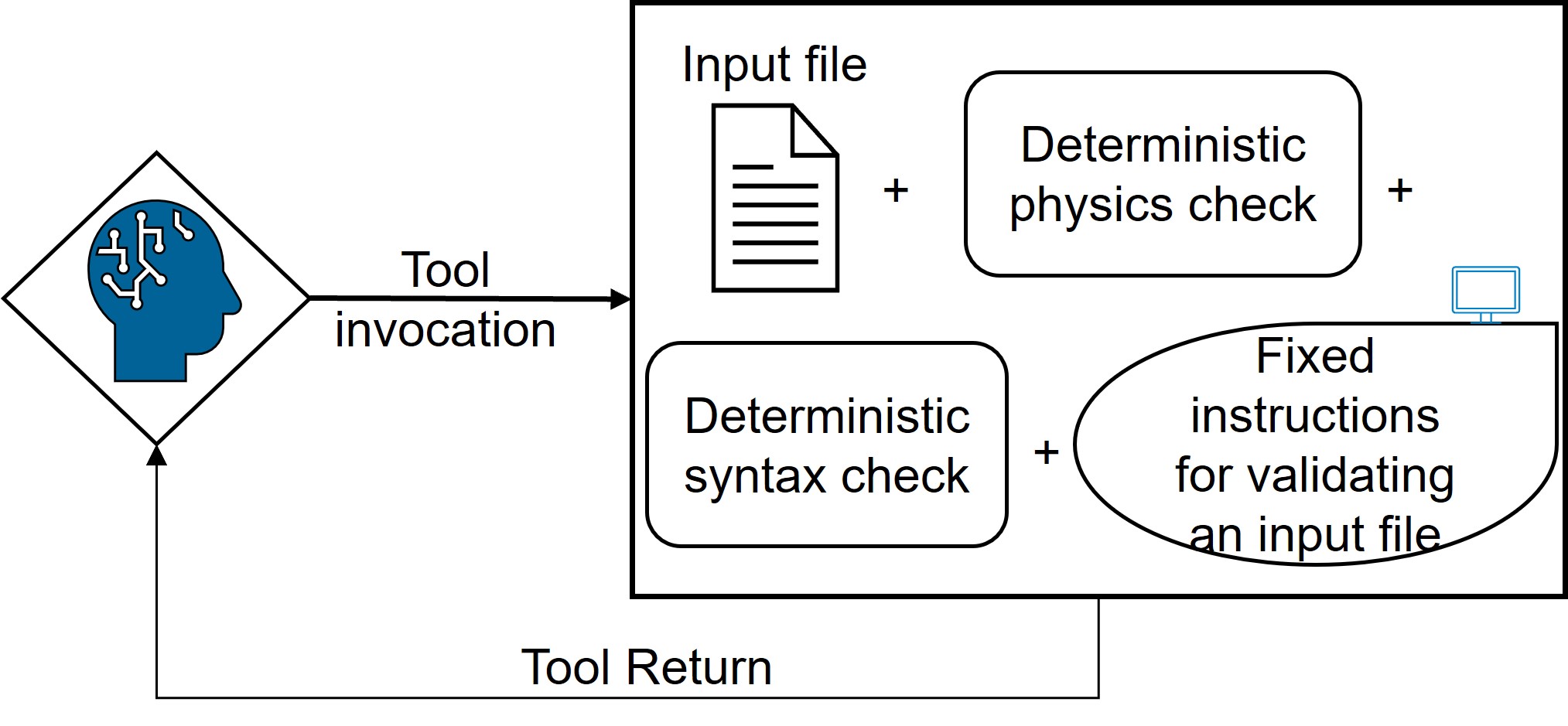}
        \caption*{(b)}
    \end{subfigure}
    \caption{Depiction of the instructions parsing tools: (a) input file creator tool and (b) input file validator tool.}
    \label{fig:tools_parsing}
\end{figure}

The input file creator tool is similar to a text file reader tool, but it has a set of expandable instructions that the developer could expand based on the experience with the agent. The tool is used to automate the generation of the solver's input file by parsing an intermediate file that was generated by the agent after processing different documents. %The idea of the intermediate file is to provide the user with intermediate step to verify, add, and edit the data gathered by the agent from different resources. 
The tool passes the intermediate file structured content to the agent. The core output of this tool is a detailed LLM prompt that instructs the AI to generate a complete and valid input file. This prompt includes explicit instructions to infer and synthesize all missing blocks, functions, boundary conditions, and solver parameters using engineering best practices. The prompt mandates that a proper title, brief problem description, and internal comments be included in the final file. The resulting file is intended to be fully executable without human intervention. 

The SAM input file validator tool is designed to verify the completeness and structural integrity of an input file. It performs both deterministic physics verification, deterministic syntax check, and provides additional instructions to the agent for verifying an input file. The tool begins by reading the file contents and checking for the presence of all required input blocks. Each block is verified, and missing entries are reported to the agent. The physics check could include units verification, geometry connectivity, functions and values check, and initial and boundary conditions. The tool also provides general semantic instructions other than the deterministic ones.

\section{Results}
\label{sec:results}

To demonstrate the capabilities of the proposed agent, four test cases of increasing complexity are considered. All cases were previously unseen by the agent and are selected to systematically evaluate its ability to construct executable system-code input files under progressively more demanding modeling conditions. The test suite includes: (i) a single pipe model, (ii) a solid-fuel system with temperature reactivity feedback, (iii) the reactor core of the Advanced Burner Test Reactor (ABTR), and (iv) the primary loop of the Molten Salt Reactor Experiment (MSRE).

The first two cases rely exclusively on structured input data provided in spreadsheet (Excel) format. These cases are designed to assess the agent’s baseline capability to generate valid input files when all required information is explicitly available. The single-pipe case evaluates the agent’s ability to model a fundamental physical component governed by one-dimensional conservation equations, while the second case introduces additional coupled physics and examines the agent’s ability to incorporate and represent physics feedback mechanisms.

The third and fourth cases move beyond structured inputs and represent realistic engineering applications with substantially higher complexity. These cases require the agent to extract, interpret, and reconcile information from multi-modal sources, including PDF documents, images, and spreadsheets. They are used to assess the agent’s ability to construct coherent, system-level models for real reactor configurations, where information is distributed across heterogeneous documents and modeling decisions involve increased structural and physical complexity.

\subsection{Test case 1: Pipe model using structured document}
\label{subsec:testcase1}
This test case evaluates whether the agent can correctly translate a fully specified, structured dataset into an executable system-code input file for a single pipe. The objective is to verify that the agent can build the appropriate code objects for a one-dimensional flow component and populate them consistently from tabular inputs. Second, it serves as a baseline physics check, since the governing behavior is dominated by the fundamental one-dimensional conservation equations of mass, momentum, and energy that form the core of system thermal--hydraulic solvers.

The problem is a steady-state simulation of a one-meter, vertical sodium pipe with an adiabatic outer wall. The boundary conditions consist of a prescribed inlet velocity and inlet temperature, together with a fixed outlet pressure, resulting in a single-phase forced-convection flow. From a modeling perspective, this requires the agent to define the pipe geometry and flow orientation,specify thermophysical properties (e.g., sodium as the working fluid), select the spatial and time numerical discretizations, apply compatible inlet/outlet boundary conditions that yield a well-posed steady-state solution, and assign numerical discretizations, solver settings and initial conditions. 

This case is intentionally designed so that the expected solution is physically easily interpretable. Because the outer wall is adiabatic and no volumetric heating is applied, the control-volume energy balance implies that the fluid experiences no net heat addition or removal along the pipe. Under these conditions, the steady-state solution should converge to an axially uniform temperature field: the bulk fluid temperature remains constant along the pipe and the wall temperature remains equal to the fluid temperature everywhere. Therefore, the bulk outlet temperature should equal the prescribed inlet temperature, and the average wall temperature should match the inlet temperature as well.

In the results, these expectations provide straightforward validation criteria for both the generated input file and the resulting simulation: a constant axial temperature profile, agreement between inlet and outlet bulk temperatures, consistency between fluid and wall temperatures, the pressure decreases monotonically in the flow direction due to frictional losses.

\begin{figure}[H]
    \centering
\includegraphics[width=1\textwidth]{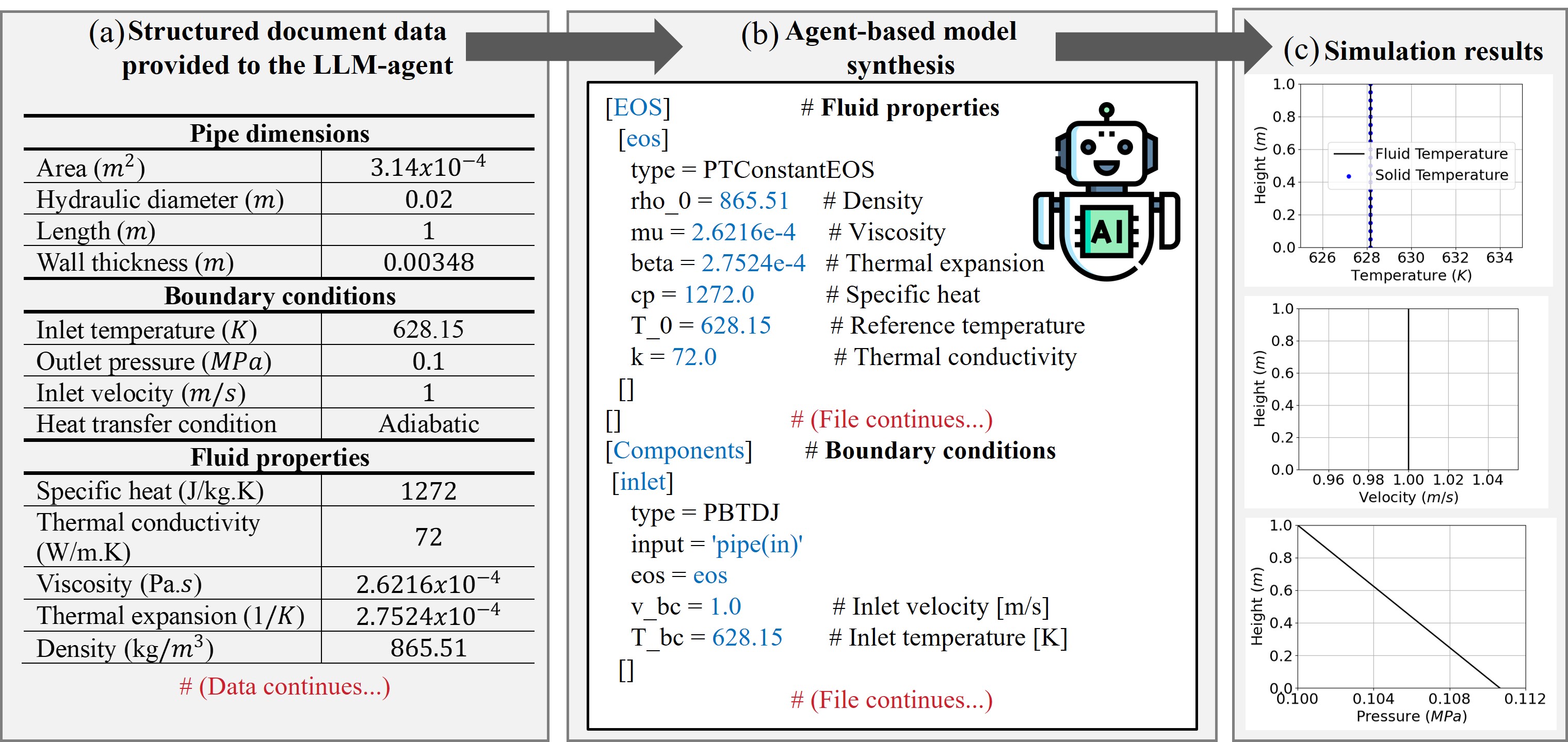} % Change to your file name and extension
    \caption{Test case 1: (a) structured document input to the LLM agent, (b) agent-generated simulation model, and (c) corresponding simulation results.}
    \label{fig:testcase1}
\end{figure}

Figure \ref{fig:testcase1} shows the workflow where structured engineering data are provided directly to the LLM-based agent. In panel (a), the inputs are already organized in spreadsheet (pipe dimensions, boundary conditions, and fluid properties), so the agent does not need to interpret diagrams or narrative text. Using these structured values, the agent automatically constructs the simulation input file in panel (b), translating the provided geometry, material/fluid properties, and boundary conditions into a complete, runnable model specification.

Panel (c) presents the corresponding simulation outputs. The agent produced a physically consistent model, applied boundary conditions correctly, and preserved conservation behavior. The panels demonstrate that when clean structured data are supplied, the agent can reliably synthesize a valid input model and produce results consistent with engineering expectations.

\subsection{Test case 2: Solid-fuel temperature reactivity feedback using structured documents}

This test case extends the baseline one-dimensional flow modeling capability demonstrated in Test Case 1 by introducing additional coupled physics in the form of temperature-driven reactivity feedback. The objective is to evaluate whether the LLM-based agent can correctly incorporate physics models that depend on the interaction between fluid flow, solid heat transfer, and system feedback mechanisms, rather than modeling fluid conservation equations alone.

The modeled system consists of a single heated flow channel with concentric solid regions representing nuclear fuel and surrounding cladding. In addition to the one-dimensional fluid-flow equations governing the coolant channel, the model includes heat conduction within the solid regions and a temperature-dependent reactivity feedback model. Internal volumetric power generation is set to zero, such that changes in solid temperature arise solely from externally imposed boundary conditions rather than internal heat sources. This design isolates the feedback mechanism and simplifies physical interpretation of the results.

A transient inlet boundary condition is applied to drive the system response. Specifically, the inlet coolant temperature is increased linearly from 628.15~K to 728.15~K over a 10~s interval. This temperature ramp propagates through the coolant channel and into the solid regions via convective heat transfer, producing a corresponding increase in fuel temperature. A negative temperature-feedback coefficient is prescribed, representing a stabilizing feedback mechanism in which increasing fuel temperature reduces system reactivity.

From a physical standpoint, the expected system behavior is straightforward and well defined. As the inlet temperature increases monotonically, the coolant temperature throughout the channel rises, leading to a gradual increase in the fuel temperature. Because the feedback coefficient is negative, the associated reactivity feedback should decrease smoothly and monotonically over time. Once the inlet-temperature ramp terminates, both the fuel temperature and the reactivity feedback are expected to asymptotically approach steady values.

\begin{figure}[H]
    \centering
\includegraphics[width=1\textwidth]{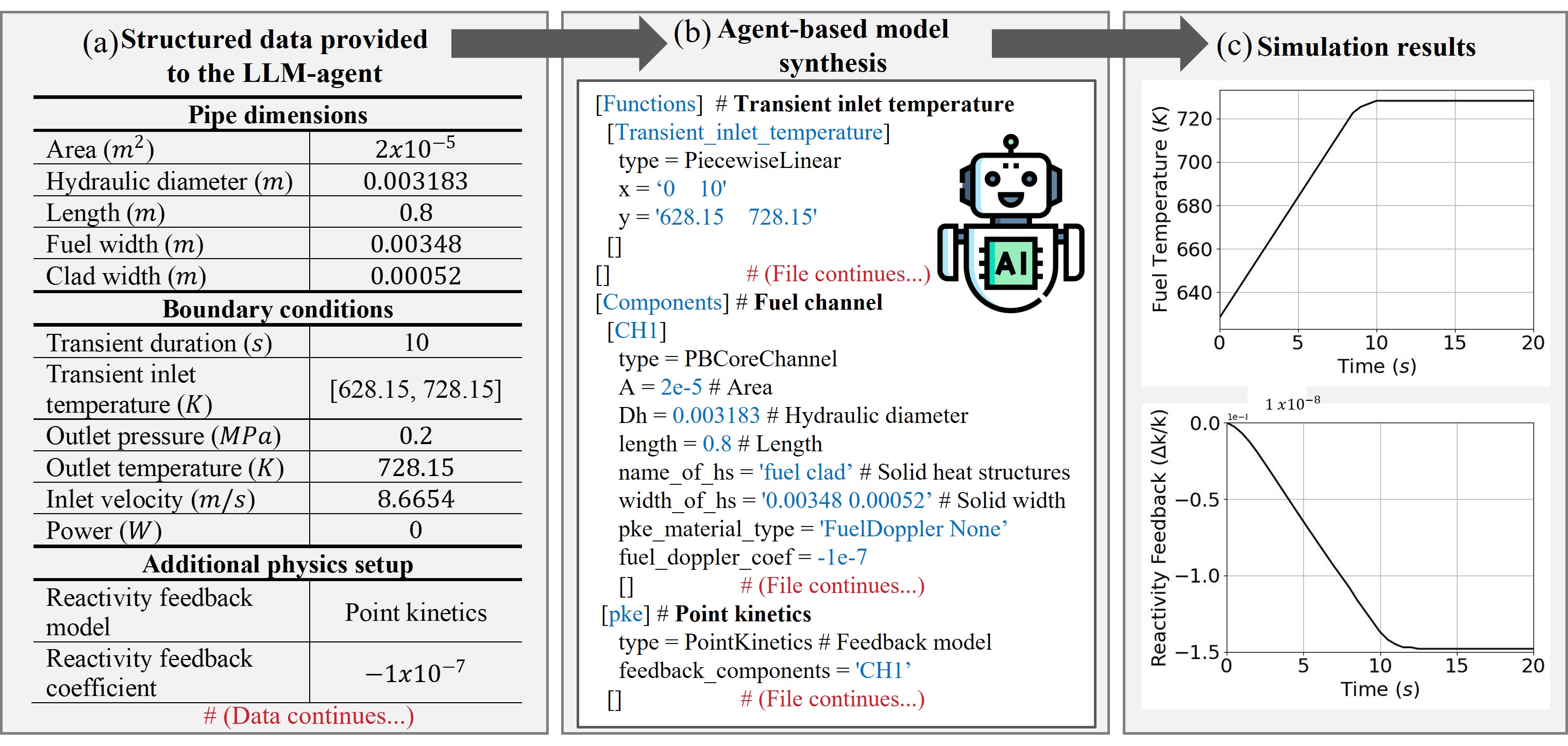} % Change to your file name and extension
    \caption{Test case 3: (a) unstructured document input to the LLM agent, (b) agent-generated structured data, (c) agent-generated model/solver's input file, and (d) corresponding simulation results.}
    \label{fig:testcase2}
\end{figure}

Figure~\ref{fig:testcase2} summarizes the workflow and results for this test case. Panel~(a) shows the structured input data provided to the agent, including tables defining the channel geometry, material properties, transient inlet-temperature boundary condition, and the temperature-feedback model parameters. Panel~(b) illustrates the agent-generated simulation input file, which includes a time-dependent inlet-temperature function, the coupled fluid--solid channel definition, and the configuration of the reactivity feedback model. Panel~(c) presents the resulting transient responses: the fuel temperature follows the imposed inlet-temperature ramp and levels off after the ramp concludes, while the reactivity feedback becomes progressively more negative over time, consistent with the specified feedback coefficient and the monotonic temperature increase.

These results demonstrate that the agent can correctly integrate additional physics models into the generated input deck, define transient boundary conditions, and produce simulations that exhibit physically consistent coupled behavior. Compared to Test Case 1, this case verifies the agent’s ability to move beyond single-physics component modeling and accurately represent simplified multiphysics interactions driven by structured input data.

\subsection{Test case 3: Advanced Burner Test Reactor (ABTR) core model}

This test case transitions from structured, fully specified inputs (Test Cases 1--2) to a more realistic engineering scenario where key modeling information is distributed across unstructured and multi-modal sources. The goal is to evaluate whether the LLM-based agent can interpret a system layout from an image, extract operating conditions and material/physics details from a PDF, reconcile these heterogeneous inputs into a consistent structured representation, and generate an executable transient model that preserves the intended topology and physical behavior.

The Advanced Burner Test Reactor (ABTR) is a conceptual sodium-cooled fast reactor developed as a test and demonstration platform for advanced fuel cycles and fast-spectrum reactor technologies~\cite{chang2008advanced}. In ABTR-type designs, liquid sodium serves as the primary coolant and flows through multiple parallel fuel subassemblies in the reactor core, where heat generated in the fuel is transferred to the coolant before being transported to downstream components.

The modeled system represents a simplified liquid-metal coolant flow path through an ABTR core configuration. Liquid sodium enters an upstream pipe and passes through a junction that splits the flow into five parallel vertical channels, representing multiple core subchannels. Downstream, the five channels recombine through a second junction into a single outlet pipe that discharges to a fixed-pressure boundary. At the inlet, velocity and temperature are prescribed (3.25~m/s and 628~K, respectively), while the outlet pressure is fixed at 0.1~MPa to close the hydraulic boundary conditions. This configuration is representative of a common reactor-core flow pattern: distribution through parallel heated channels followed by mixing and discharge.

Thermal energy is added through the heated channel region, with a total heat input of 250~MW distributed among the five channels. Each channel includes a simplified solid region surrounding the coolant (e.g., fuel and cladding), enabling a basic representation of conjugate heat transfer: heat is deposited in the solid, conducted through surrounding layers, and transferred to the coolant via convection. As a result, the coolant temperature increases along the channel axial direction, and the outlet temperature depends on both the power distribution and the channel flow split. 

The expected system response provides multiple physics-based validation checks. First, an uneven flow distribution among the five parallel channels is anticipated because junction geometry and branch-specific hydraulic resistances are not identical, leading to different pressure drops across each branch. Second, pressure should decrease along the flow direction due to frictional losses in the pipes/channels and additional form losses at the splitting and merging junctions. Third, coolant temperatures should rise through the heated channels. 

\begin{figure}[H]
    \centering
\includegraphics[width=1\textwidth]{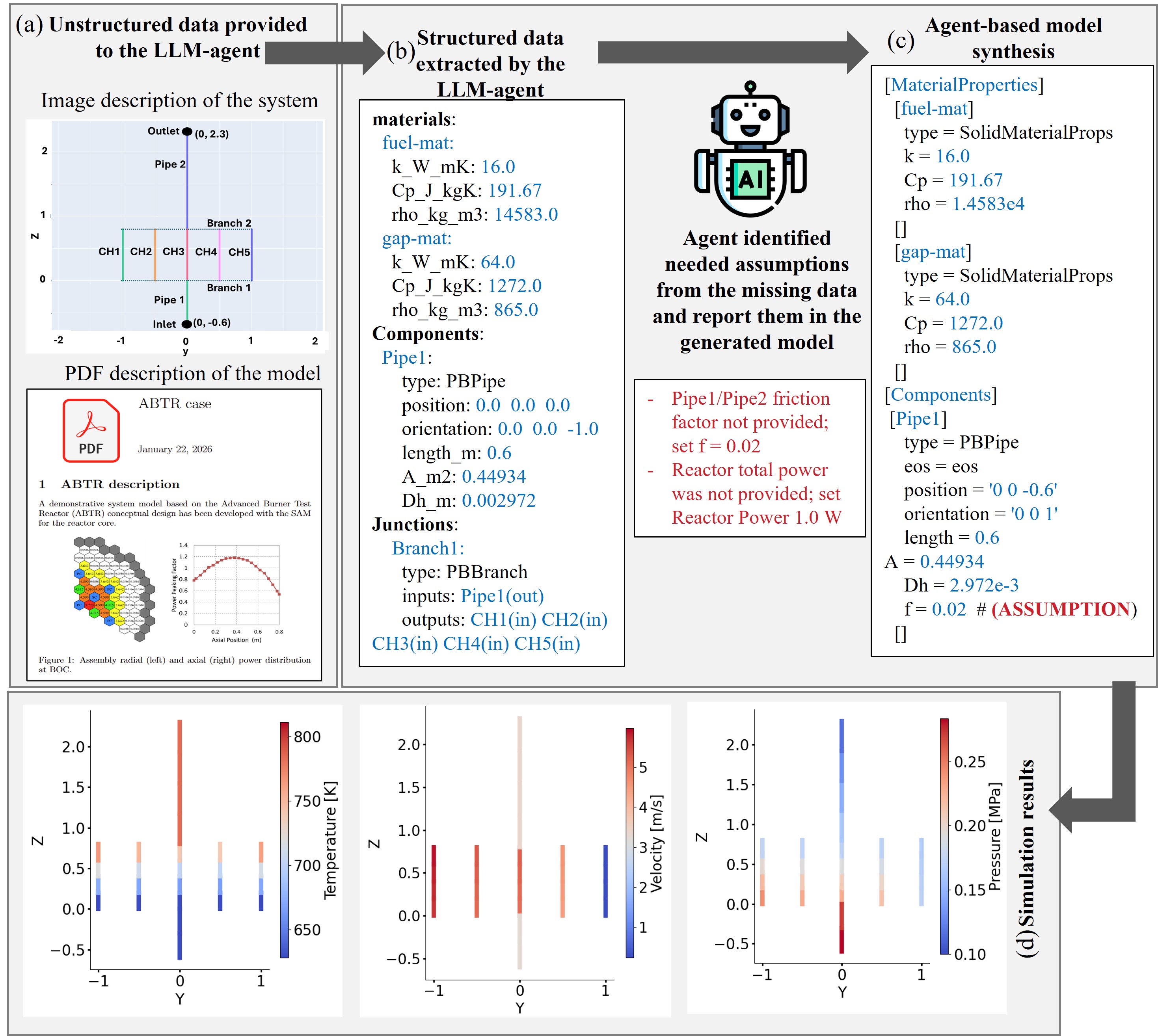} % Change to your file name and extension
    \caption{Test case 3: (a) structured document input to the LLM agent, (b) agent-generated simulation model, and (c) corresponding simulation results.}
    \label{fig:testcase3}
\end{figure}

Figure~\ref{fig:testcase3} illustrates the end-to-end workflow for this case. In panel~(a), the agent ingests unstructured multi-modal inputs: an image that encodes the system topology (components and connectivity) and a PDF containing case information such as boundary conditions, material properties, and operating parameters. In panel~(b), the agent converts these sources into a structured intermediate representation (e.g., a component-and-connection table with extracted parameter values) that can be reviewed and edited by a user if necessary. Using this structured representation, the agent then generates the final system-code input/model. When required details are missing or ambiguous in the source documents, the agent explicitly identifies the gaps and records any substituted values as assumptions to preserve transparency and traceability. Finally, panel~(d) presents representative simulation outputs (e.g., temperature, velocity, and pressure fields), which exhibit trends consistent with the expected thermal--hydraulic behavior of a parallel-channel heated core segment.

\subsection{Test case 4: Molten Salt Reactor Experiment (MSRE) primary loop model}
\label{subsec:testcase4}

\begin{figure}[H]
    \centering
\includegraphics[width=1\textwidth]{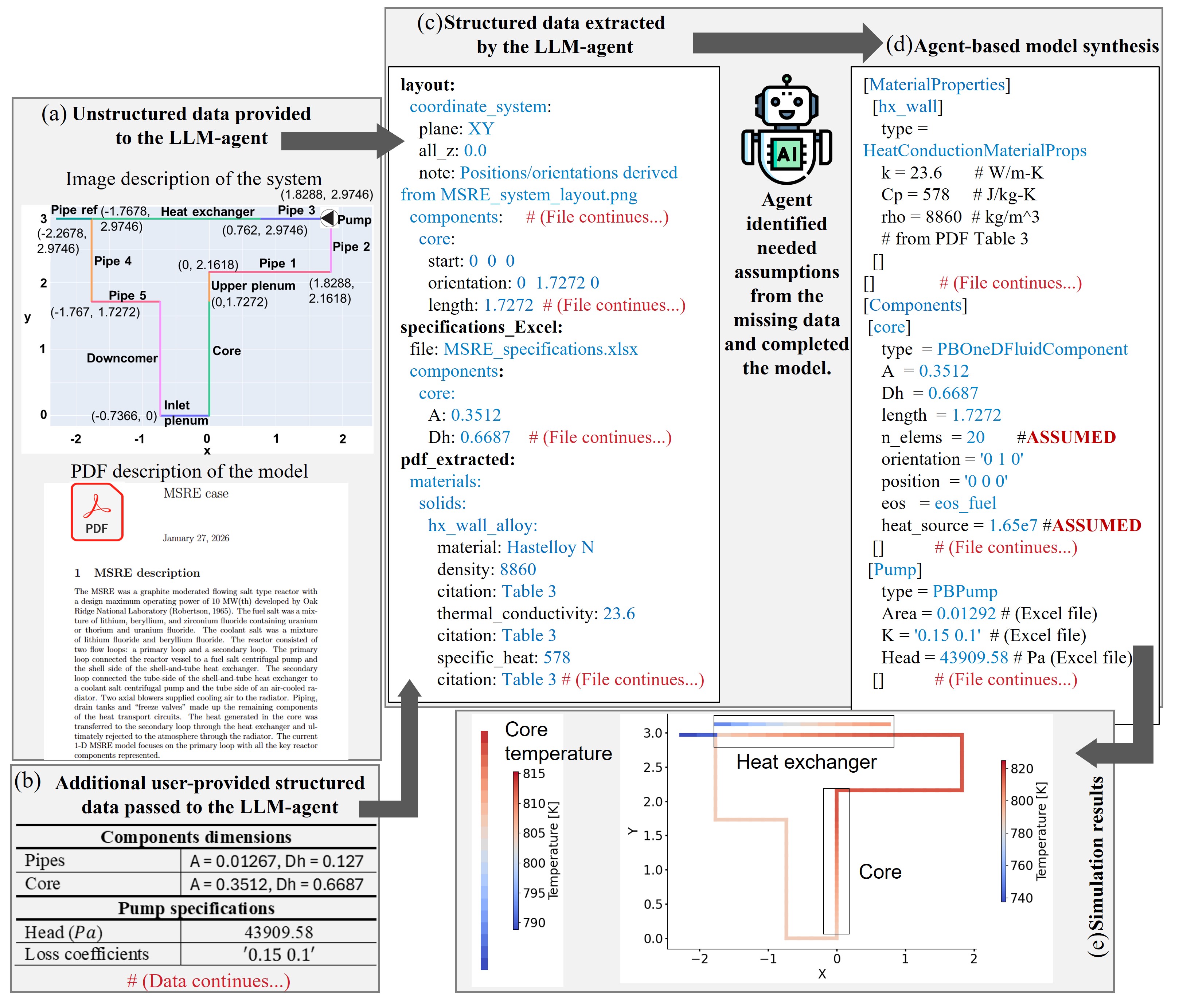}
    \caption{Test case 4: (a) multi-modal document inputs to the LLM agent, (b) agent-generated simulation model/input file, and (c) corresponding simulation results.}
    \label{fig:testcase4}
\end{figure}

This test case evaluates the agent’s ability to build a system-level circulating-loop model representative of real engineering practice, where the topology spans multiple components (core, plenums, pump, heat exchanger, and connecting piping) and required information is distributed across heterogeneous sources. The modeled system is based on the primary loop of the Molten Salt Reactor Experiment (MSRE)~\cite{haubenreich1970experience}, a closed-loop molten-salt system in which a heated section adds thermal energy to the primary salt and a heat exchanger transfers that energy to a secondary coolant circuit while a pump maintains circulation. Beyond demonstrating correct component-level modeling, this case specifically tests whether the agent can reconstruct connectivity from an unstructured layout, reconcile parameters extracted from PDFs and images with user-provided structured data, and synthesize an executable input deck that produces physically consistent loop-wide trends.

The primary loop flows through an inlet plenum, a heated core segment, an upper plenum, connecting pipes, a pump, and the primary side of a heat exchanger, before returning through additional piping and a downcomer to close the loop. The secondary (coolant-salt) side of the heat exchanger is represented using inlet/outlet boundary conditions, which emulate an external cooling circuit without explicitly modeling the full secondary loop. A volumetric heat source is applied in the core region, raising the primary-salt temperature as it passes through the heated section. Heat is then removed in the heat exchanger as energy is transferred to the secondary side, producing a temperature decrease in the primary loop across the exchanger and a corresponding temperature increase in the secondary coolant from its inlet to outlet.

The expected behavior provides intuitive validation criteria commonly used for circulating thermal-hydraulic loops. First, the hottest temperatures should occur near the core outlet, where heat addition is completed. Second, a measurable temperature drop should occur across the heat exchanger on the primary side, with the magnitude governed by the imposed secondary boundary conditions and exchanger specifications. Third, the secondary side should exhibit a temperature rise from inlet to outlet consistent with net heat gain. Finally, after transient effects diminish, the system should settle into a self-consistent loop temperature profile (hot leg leaving the core and cooling leg after the heat exchanger/return path), indicating that mass and energy balances are satisfied around the loop.

Figure~\ref{fig:testcase4} summarizes the end-to-end workflow for this case. In panel~(a), the agent ingests multi-modal inputs: an image encoding the loop layout and connectivity (core, plenums, pipes, pump, and heat exchanger) and a PDF providing case descriptions, material definitions, and operating/boundary conditions. In panel~(b), the agent extracts these details and converts them into an intermediate structured representation (e.g., a component list, connectivity graph, and parameter tables) that can be reviewed and supplemented. In this case, additional user-provided structured data (such as component dimensions and pump specifications) are incorporated to resolve parameters that are missing or ambiguous in the unstructured sources. Using the combined dataset, the agent synthesizes a complete simulation input model, explicitly flagging any remaining gaps as assumptions when required for executability and traceability. Panel~(e) shows representative simulation results: temperatures increase through the heated core region and decrease across the heat exchanger on the primary side, while the secondary side warms across the exchanger. These trends are consistent with expectations for a pumped heated loop with heat removal, demonstrating that the generated model is executable and produces physically reasonable loop-level behavior.

\subsection{Results analysis}
\label{subsec:analysis}

\begin{figure}[H]
    \centering
\includegraphics[width=1\textwidth]{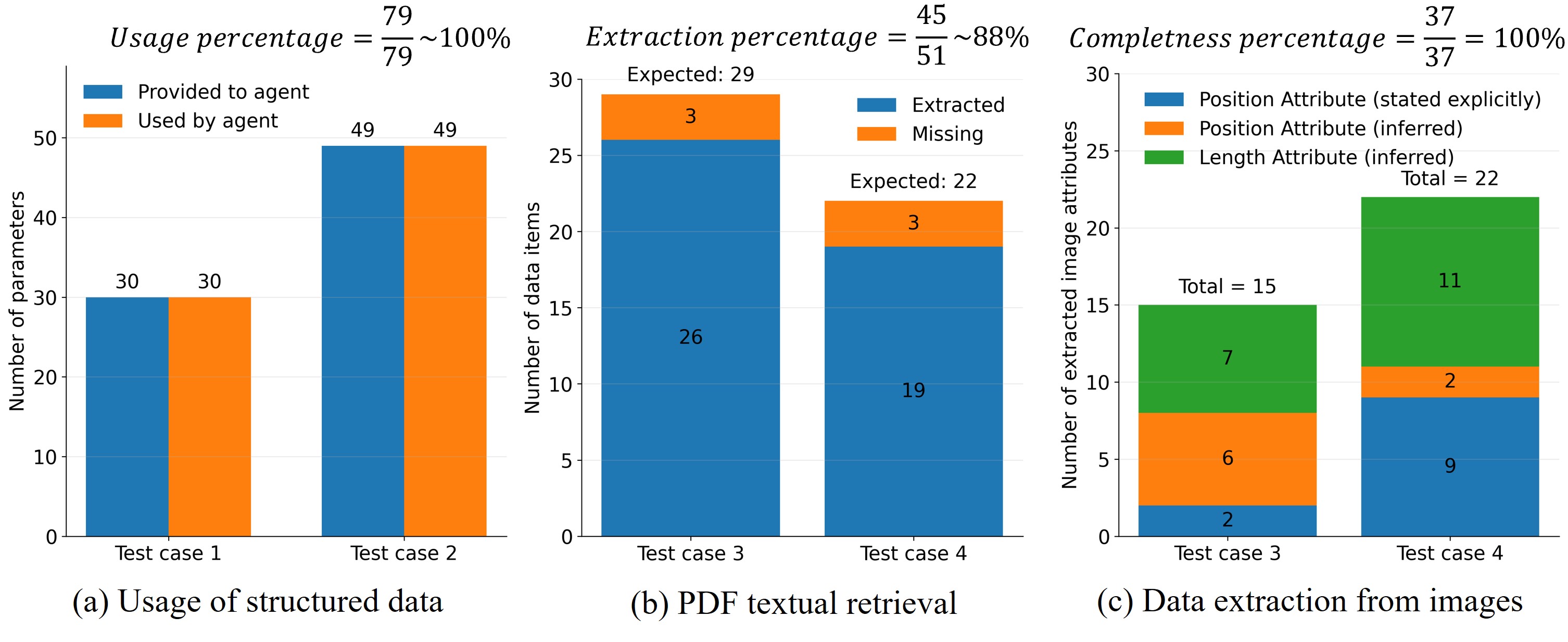} % Change to your file name and extension
    \caption{Performance of the agent’s data pipeline across structured inputs, PDF text retrieval, and vision-based image extraction. (a) All structured parameters passed to the agent are utilized, yielding a usage rate of 100\%. (b) PDF textual retrieval recovers 45 of 51 expected items. (c) Vision-based extraction achieves 100\% completeness, breaking extracted image attributes into explicitly stated position, inferred position, and inferred/deduced length components.}
    \label{fig:quality}
\end{figure}

Figure~\ref{fig:quality} provides a quantitative summary of how information is consumed, extracted, and propagated through the agent’s multi-modal data pipeline across the four test cases. The purpose of this analysis is to contextualize the results presented in Sections~\ref{subsec:testcase1}--\ref{subsec:testcase4} by examining how different data sources contribute to the final models. Importantly, this analysis is intended to characterize information flow and coverage within the pipeline rather than to serve as a formal quality or performance assessment of the agent.

Panel~(a) focuses on structured data usage for Test Cases~1 and~2, where all required inputs are provided in spreadsheet form. The bars compare the number of parameters supplied to the agent with the number ultimately used in the generated simulation inputs. In both cases, all provided parameters (30 for Test Case~1 and 49 for Test Case~2) are fully consumed, resulting in a 100\% usage rate. This confirms that no structured information is dropped or ignored during ingestion and translation, and that the agent preserves all explicitly supplied data when generating input files.

Panel~(b) examines PDF-based textual retrieval performance for Test Cases~3 and~4 by comparing the number of expected data items against those successfully extracted by the agent. Across both cases, the agent retrieves the majority of the required textual information (45 out of 51 total items), corresponding to an overall extraction rate of approximately 88\%. While these omissions do not prevent model execution, they highlight where human review or supplemental structured inputs may be beneficial in complex, document-driven scenarios.

Panel~(c) reports performance for image-based information extraction in Test Cases~3 and~4. All required image-derived attributes (37 out of 37) are successfully recovered, yielding a completeness rate of 100\%. The bars further decompose the extracted attributes into three categories: explicitly stated positional information, inferred positional attributes, and inferred or deduced length-related attributes. This breakdown illustrates that complete recovery often requires reasoning beyond direct visual cues, particularly when dimensions or relative placements must be inferred from layout diagrams rather than explicitly labeled. The results demonstrate that the agent can not only parse visual structure, but also infer missing geometric relationships needed for model construction.

\begin{figure}[H]
    \centering
\includegraphics[width=1\textwidth]{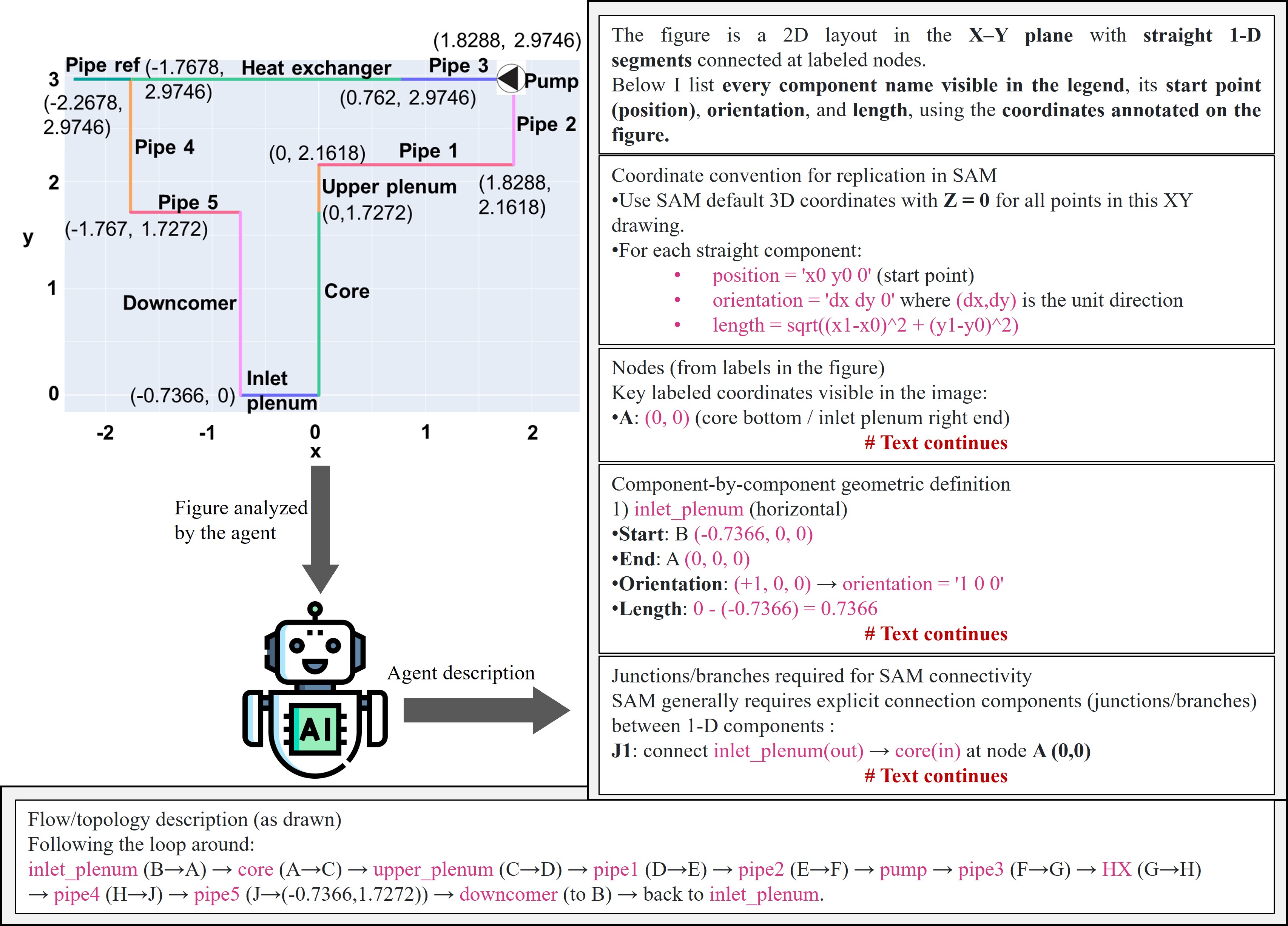} % Change to your file name and extension
    \caption{Agent-assisted extraction of 1-D SAM component geometry and connectivity from a 2-D loop schematic (Test Case~4). The agent identifies labeled nodes and coordinates, defines each component by start point, orientation, and length, specifies required junction/branch connectivity, and summarizes the loop flow path for consistency checking.}
    \label{fig:image_ex}
\end{figure}

Figure~\ref{fig:image_ex} illustrates how the agent converts the Test Case~4 2-D X--Y schematic into a 1-D, solver-ready geometric and topological specification suitable for solver. The schematic (left) depicts the primary loop as a sequence of straight connected segments labeled \emph{inlet plenum}, \emph{core}, \emph{upper plenum}, \emph{pipe 1--5}, \emph{pump}, and \emph{heat exchanger}. Key node coordinates are annotated at corners and endpoints, providing a minimal but sufficient geometric details for reconstructing component start/end locations and connectivity.

The agent’s extracted interpretation is demonstrated on the right. First, it establishes a coordinate convention for replication in the solver. It then enumerates labeled nodes from the figure (e.g., node identifiers and their coordinates) and uses these nodes to define each 1-D component segment by a start point, an orientation vector aligned with the segment direction, and a length computed from the annotated coordinates. This component-by-component decomposition is critical for system-code model construction because SAM requires explicit geometric definitions for each 1-D element.

In addition to geometry, the agent explicitly reconstructs topology. Because SAM generally requires connection elements (e.g., junctions/branches) to link 1-D components, the agent lists the required junction definitions and maps component ports (inlet/outlet) to shared nodes. Finally, a concise flow/topology summary line provides an end-to-end traversal of the loop (from inlet plenum through the core, pump, and heat exchanger and back through the downcomer), serving as a high-level consistency check that the reconstructed graph matches the intended physical flow path. Overall, the figure highlights the agent’s ability to extract actionable geometric and connectivity information from a schematic, producing intermediate structured definitions that are both human-verifiable and directly usable for automated input-deck generation.

\section{Discussions}
\label{sec:discuss}
\subsection{Generalization of the methodology}

The proposed methodology functions as a generalized framework applicable across various physics solvers and simulation codes. While the core architecture remains consistent, specific domain adaptation is necessary to address unique solver requirements, such as input file syntax, physics constraints, and parameter definitions. This process requires a modular update of system instructions and the embedding of relevant documentation.

Crucially, the agentic tools are categorized by their dependence on the solver:

\begin{itemize}
    \item \textbf{Solver-Specific Tools:} Tools like the \textit{Input File Creator} require adaptation to match the target solver's syntax. For example, in the SAM implementation, this involved adding deterministic calculations and checks to the input file validator.
    \item \textbf{Solver-Agnostic Tools:} General-purpose utilities, such as the Excel file reader, text file reader, Python execution tool, and RAG implementation remain unchanged regardless of the simulation environment.
\end{itemize}

\subsection{Privacy concerns}
Incorporating LLMs into scientific workflows raises important privacy and data protection concerns, particularly when dealing with sensitive domains such as nuclear engineering. One critical distinction is between using publicly hosted chatbots (e.g., via web interfaces) versus interacting with LLMs through private APIs or locally hosted models. Online chatbots typically route user data through external servers, where prompts and uploaded documents may be logged, used for model improvement, or subject to data retention policies outside the user's control. This is particularly concerning when working with proprietary models, internal workflows, or classified datasets. By contrast, using LLMs through secured APIs or deploying them on local infrastructure allows users to retain full control over their data. For instance, documents such as reactor design specifications, experimental results, or safety analyses can be processed without risk of data leakage. When RAG is employed, vector databases and embedding models can be hosted on secure servers, ensuring that the retrieval pipeline does not expose sensitive content externally.

\subsection{Safety concerns}
Beyond privacy, the safe use of LLM-powered agents in nuclear applications requires careful consideration. SAM is a system-level thermal-hydraulics code developed and used in regulatory licensing activities by the U.S. Nuclear Regulatory Commission (NRC). Given the critical nature of the simulations it supports, introducing automation via LLMs must not compromise the integrity or reliability of the input files. To address these safety concerns, we introduce a layered safety mechanism. First, the agent does not directly write the final SAM input file upon extracting data. Instead, it generates an intermediate structured file containing all the extracted parameters and source references. This intermediate file allows human users to verify, edit, and approve the content before the SAM input file is finalized. Second, the modular prompt design ensures transparency and traceability in tool invocation, making it easier for users to follow the logic of the agent’s actions. While the agent significantly streamlines the workflow, ultimate responsibility for correctness and safety remains with the user. Manual review is essential, particularly in high-stakes environments like nuclear licensing. By combining automation with human oversight, the system balances efficiency and safety, minimizing the risk of unverified input generation while still benefiting from AI-driven acceleration.

\subsection{Model limitations}
Despite their impressive capabilities, LLMs are not without limitations—especially when applied to scientific and safety-critical domains. One of the primary concerns is the potential for hallucinations, where the model generates information that is plausible-sounding but factually incorrect. To mitigate these risks, the agent described in this work employs a hybrid strategy that combines tool chaining, rule-based validation, and human oversight. Tool chaining ensures that domain-specific calculations are delegated to deterministic tools rather than relying only on the LLM’s internal reasoning. Rule-based validation steps are integrated into the workflow to detect physically implausible values or missing parameters. Most importantly, a human-in-the-loop review is enforced by generating an intermediate file before finalizing the SAM input file. This gives users the opportunity to audit the output and correct any inconsistencies.

Despite these limitations, the agent demonstrated strong potential in supporting domain-specific problem-solving. In the ABTR test case, the model successfully diagnosed and addressed a user concern related to high system pressure. It traced the issue to an incorrect form loss value, recommended a correction, and updated the input accordingly. This case illustrates the LLM’s potential to reason through physical problems when equipped with the right tools and context. However, the success also highlights the importance of maintaining a collaborative approach where AI augments expert knowledge rather than replacing it. In this work, code execution, error reporting to the agent, and data analysis were performed manually by the user. Subsequent efforts have focused on automating these workflow tasks, including code execution and data analysis. Furthermore, an error-resolution tool and an accompanying error database have been developed to support the agent’s operations.

%This multi-stage process ensures that even non-programmatic users can contribute to simulation development while retaining full control and traceability over the simulation input.
\subsection{Impact}

The primary impact of this work is the demonstration that multi-modal, tool-augmented LLM agents can directly reduce the "model setup bottleneck" that dominates many engineering simulation workflows where model construction depends on data spread across heterogeneous documents. By converting mixed source materials (PDFs, images, spreadsheets, and text) into a human-auditable intermediate structured representation, the framework enables systematic synthesis of solver-ready input decks. This supports a broader paradigm in which “modeling becomes prompting,” i.e., users communicate intent and provide supporting documentation, and the agent translates that intent into transparent, executable simulation models. A second, longer-term impact is how this approach could reshape the future of modeling and simulation practice. Instead of treating input decks as one-off artifacts created by hand, models become regenerable products of a document-grounded pipeline: as designs evolve and documentation changes, the model can be rebuilt systematically with traceable provenance to source material and explicit assumption labeling.

For experienced analysts, the AutoSAM functions as a productivity assistant. It can (i) automate model assembly, (ii) reduce time spent searching manuals and transcribing requirements by enabling retrieval over both text and figures, and (iii) surface gaps early by explicitly identifying missing parameters and flagging assumptions, allowing experts to focus on model fidelity, scenario definition, and interpretation. Critically, the intermediate structured file preserves expert control by serving as a deliberate checkpoint for review and correction before final input generation. 

\section{Conclusions}
\label{sec:conclusions}

In this paper, we presented AutoSAM, a multi-modal, retrieval-augmented LLM agent framework for automating the generation of input files for the SAM system thermal-hydraulics code. The central contribution is a unified workflow that enables a single ReAct-based agent to ingest heterogeneous engineering documentation, including reactor design reports, system diagrams, material property tables, and spreadsheets, and produce validated, solver-compatible SAM input decks through an auditable, two-stage process. By combining SAM-specific system instructions, retrieval-augmented generation over the code's user guide and theory manual, and seven specialized tools for document parsing, input creation, and input validation, the framework addresses the well-recognized bottleneck of manual input-file preparation that currently constrains reactor modeling productivity.

Four case studies demonstrated the framework's capabilities across a range of modeling complexity representative of practical SAM applications. The single-pipe case confirmed that the agent correctly constructs basic one-dimensional flow components and applies boundary conditions that yield physically consistent steady-state solutions. The temperature reactivity feedback case verified the agent's ability to incorporate coupled multi-physics models, including solid heat conduction and point kinetics, into the generated input deck. The ABTR core case demonstrated multi-modal document processing: the agent extracted system topology from an engineering diagram, reconciled operating parameters from a PDF design document, and synthesized a five-channel parallel-flow model with conjugate heat transfer, explicitly identifying and labeling parameters that were assumed due to missing source data. The MSRE primary loop case extended these capabilities to a full circulating system with multiple component types, testing the agent's ability to reconstruct loop connectivity and produce simulations exhibiting the expected hot-leg/cold-leg temperature distribution. 

Quantitative analysis of the data pipeline revealed that all structured parameters provided via spreadsheets were utilized without loss (100\% usage rate), PDF-based text extraction achieved approximately 88\% recall of expected data items, and vision-based geometric extraction from engineering diagrams achieved 100\% completeness across 37 extracted attributes. These metrics provide an initial characterization of the pipeline's information fidelity, though systematic benchmarking against human analyst performance remains an important direction for future work. 

The results demonstrate a practical path toward what we term prompt-driven reactor modeling: a paradigm in which analysts provide system descriptions and supporting documentation, and an intelligent agent translates that intent into transparent, executable, and reproducible SAM simulations. The intermediate structured representation ensures that human expertise remains central to the process by providing an explicit review checkpoint consistent with the quality assurance practices expected in nuclear safety analysis. For experienced analysts, the framework functions as a productivity tool that automates data extraction and input assembly while preserving expert control over modeling decisions.

Despite these advances, several limitations of the current framework should be noted.  First, we have demonstrated the approach exclusively with SAM. While the architecture is designed to be solver-agnostic through modular system instructions and tool definitions, adaptation to other system codes (e.g., RELAP5, TRACE) requires re-engineering of the solver-specific components; Additionally, context window limitations may constrain the framework's applicability to very large system models with hundreds of components. Multi-agent decompositions (e.g., planner–builder–verifier roles) may improve robustness and scalability for large systems and multi-physics coupling~\cite{zhou2025autonomous}. 

Several directions for future work follow naturally from this study. An ongoing effort is extending the framework to full workflow automation, including simulation execution, iterative runtime error resolution, automated post-processing, and results reporting, would close the loop between model creation and model use. Other key research directions include multi-agent architectures for decomposing large system models, formal verification and validation protocols suitable for regulatory acceptance, extension to additional system codes and multi-physics platforms, and systematic evaluation of input-file quality against human-generated reference models. As advanced reactor designs progress toward licensing, AI-assisted modeling frameworks such as AutoSAM have the potential to accelerate safety analysis workflows while maintaining the rigor and traceability that nuclear regulation demands.

\section*{Acknowledgment}

The submitted manuscript has been co-created by UChicago Argonne, LLC, Operator of Argonne National Laboratory (“Argonne”). Argonne, a U.S. Department of Energy Office of Science laboratory, is operated under Contract No. DE-AC02-06CH11357. The U.S. Government retains for itself, and others acting on its behalf, a paid-up nonexclusive, irrevocable worldwide license in said article to reproduce, prepare derivative works, distribute copies to the public, and perform publicly and display publicly, by or on behalf of the Government.

Authors Zaid Abulawi, Zavier Ndum Ndum, and Yang Liu are partially supported by the U.S. Department of Energy Office of Nuclear Energy Distinguished Early Career Program under contract number DE-NE0009468.

\section*{Declaration of generative AI and AI-assisted technologies in the manuscript preparation process}

During the preparation of this work, the authors used generative AI tools to assist with language refinement, editing, and manuscript organization. In addition, the study itself involves the development and evaluation of an AI-based framework for automated input file generation. All AI-assisted outputs used in manuscript preparation were reviewed and edited by the authors as needed, and the authors take full responsibility for the content of the published article.

\bibliographystyle{elsarticle-num}   %% .bst file that follows ASME journal format. Do not change.

\bibliography{refs} %% <=== change this to name of your bib file

\end{document}